\pdfoutput=1

\documentclass[dvipsnames]{article}
\usepackage{arxiv}

\usepackage{amsmath,amssymb,amsthm,amsfonts,mathtools}
\usepackage{graphicx}
\usepackage{textcomp}
\usepackage{braket}
\usepackage{booktabs}
\usepackage{multirow}
\usepackage{xspace}

\usepackage{tabularx}

\newenvironment{adjustwidth}[2] %
    {}{}

\usepackage[T1]{fontenc}
\usepackage[utf8]{inputenc}

\DeclareUnicodeCharacter{0301}{\'{e}}

\usepackage{bm}

\DeclareMathAlphabet{\mathdutchcal}{U}{dutchcal}{m}{n}
\SetMathAlphabet{\mathdutchcal}{bold}{U}{dutchcal}{b}{n}
\DeclareMathAlphabet{\mathdutchbcal}{U}{dutchcal}{b}{n}
\DeclareMathAlphabet{\mathpzc}{OT1}{pzc}{m}{it}

\makeatletter
\let\MYcaption\@makecaption
\makeatother
\usepackage{subcaption} %
\makeatletter
\let\@makecaption\MYcaption
\makeatother

\usepackage{tablefootnote}

\usepackage[
    backend=biber
  ,uniquename=false
  ,giveninits=true                  %
  ,url=false, isbn=false, doi=false %
  ,sorting=none             %
  ,date=year                %
  , maxnames=4
  , maxbibnames=99
  , minnames=1
  ,style=authoryear
  ]{biblatex}
\AtEveryBibitem{                      %
  \iffieldequalstr{eprinttype}{jstor}
  {\clearfield{eprint}}
  {}
  \clearfield{urlyear}
  \clearfield{urlmonth}
  \clearfield{url}
}
\renewrobustcmd*{\bibinitdelim}{\,} %
\renewcommand{\cite}[1]{\autocite{#1}}

\usepackage{lipsum}

\usepackage[boxed,ruled,vlined,linesnumbered]{algorithm2e}
\DontPrintSemicolon
\SetKwProg{Fn}{function}{}{}
\SetKwFunction{FnSampleFree}{SampleFree}
\SetKwFunction{FnRestartArm}{RestartArm}
\SetKwFunction{FnPickArm}{PickArm}
\SetKwFunction{FnRewire}{Rewire}
\SetKwFunction{FnNearest}{Nearest}
\SetKwFunction{FnRestartArm}{RestartArm}
\SetKwComment{Comment}{$\triangleright$\ }{}
\SetKwInput{KwInit}{Initialise}

\usepackage{cleveref}                                        %
\crefname{assumption}{assumption}{assumptions}
\crefname{problem}{problem}{problems}
\crefname{algorithm}{Alg.}{Algs.}
\Crefname{algorithm}{Algorithm}{Algorithms}
\crefname{figure}{Fig.}{Figs.} %
\crefformat{equation}{(#2#1#3)}
\crefrangeformat{equation}{(#3#1#4) to~(#5#2#6)}
\crefmultiformat{equation}{(#2#1#3)}%
{ and~(#2#1#3)}{, (#2#1#3)}{ and~(#2#1#3)}

\usepackage{microtype}

\usepackage{tkz-graph}
\usetikzlibrary{shapes.multipart, positioning, decorations.markings, arrows.meta, calc}

\microtypesetup{activate={true,nocompatibility},final,tracking=true,kerning=true,factor=1100,stretch=10,shrink=10}
\usepackage{etoolbox}
\makeatletter
\pretocmd{\NAT@citexnum}{\@ifnum{\NAT@ctype>\z@}{\let\NAT@hyper@\relax}{}}{}{}
\makeatother

\newcommand*\prob[1][]{\mathbb{P}}

\usepackage{etoolbox}

\begin{document}

\newcommand{\shortheadtitle}{
  A Review of Explainable AI on Self-Attentive VisionTransformer in Health Care
  }

\title{
    Interpretable Medical Imagery Diagnosis with Self-Attentive Transformers:
    A Review of \\ Explainable AI for Health Care
  }

\author{
  Tin Lai\\
  School of Computer Science\\
  University of Sydney\\
  Australia \\
}

\maketitle

\begin{abstract}
Recent advancements in artificial intelligence (AI) have facilitated its widespread adoption in primary medical services, addressing the demand-supply imbalance in healthcare. Vision Transformers (ViT) have emerged as state-of-the-art computer vision models, benefiting from self-attention modules. However, compared to traditional machine-learning approaches, deep-learning models are complex and are often treated as a ``black box'' that can cause uncertainty regarding how they operate. Explainable Artificial Intelligence (XAI) refers to methods that explain and interpret machine learning models' inner workings and how they come to decisions, which is especially important in the medical domain to guide the healthcare decision-making process. This review summarises recent ViT advancements and interpretative approaches to understanding the decision-making process of ViT, enabling transparency in medical diagnosis applications.
\end{abstract}

\robustify\textsuperscript
\global\def\printmytablefootnoteContent{} %
\global\def\mytablefootnotecite#1{%
    \footnotemark
    \ifdefined\lockmytablefootnotecite
    \else
        \xdef\printmytablefootnoteContent{\printmytablefootnoteContent{} {\textsuperscript{\thefootnote}\cite{#1}}~}%
    \fi
}

\section{Introduction}

Artificial intelligence (AI) has made significant strides in various domains in recent years, revolutionising industries and shaping how we approach complex problems.
One of AI's most remarkable applications is in medical imaging~\cite{esteva2021deep}, where it has brought about unprecedented advancements in automated image analysis, diagnosis, and decision-making.
Medical images are one of the most common clinical diagnostic methods~\cite{shung2012principles}.
These images have properties that vary depending on the medical diagnosis and anatomical local such as skin~\cite{hu2022x,lucieri2022exaid,stieler2021skinimage,lucieri2020interpretability},
chest~\cite{lenis2020domain,hu2022x,brunese2020explainable,corizzo2021explainable,mondal2021xvitcos},
brain~\cite{bang2021spatio,li2021explainable}, liver~\cite{MOHAGHEGHI2022105106}, and others.
Deep learning algorithms have found numerous critical applications in the healthcare domain, ranging from detecting diabetes~\cite{sensorsMLforDiabetes}, genomics~\cite{Yang2021review}, and mental health support~\cite{lai2023psyllm}.
Among the latest breakthroughs in computer vision models, Vision Transformers (ViT)~\cite{dosovitskiy2020image} have emerged as cutting-edge architecture, leveraging self-attention mechanisms to achieve state-of-the-art performance in various visual tasks.

As medical professionals increasingly rely on AI-powered systems to aid in diagnosis and treatment planning~\cite{ker2017deep}, the need for interpretability and transparency in AI models becomes paramount~\cite{macdonald2022interpretable}.
Deep learning models, including ViTs, often exhibit highly complex and intricate internal representations, making it challenging for experts to comprehend their decision-making process.
The opaque nature of these models raises concerns about their reliability and safety, especially in critical applications such as medical diagnostics, where accurate and trustworthy results are of utmost importance~\cite{ghosh2020interpretable}.
\emph{Explainable Artificial Intelligence} (XAI) is a burgeoning field that seeks to bridge the gap between the black-box nature of AI algorithms and the need for understandable and interpretable decision-making processes~\cite{arrieta2020explainable}.
XAI addresses a fundamental challenge: \emph{How can we make AI's decision-making process more transparent and comprehensible to both experts and non-experts?}
While complex models might achieve impressive accuracy, their inability to provide human-readable explanations hinders their adoption in critical applications such as healthcare, finance, and legal domains. This limitation not only undermines users' trust but also poses ethical and regulatory concerns.
The integration of XAI can also lead to improved collaboration between AI systems and human experts, as well as the identification of novel patterns and insights that might have been overlooked otherwise.

In the realm of XAI, several techniques contribute to enhancing the transparency and trustworthiness of complex machine learning models.
Local Interpretable Model-Agnostic Explanations (LIME) offer insights into any model's predictions by approximating its behavior with interpretable surrogate models~\cite{ribeiro2016should}.
LIME is model-agnostic, which means it is applicable to most AI models without relying on any specific model architecture.
Gradient-based saliency methods, like Grad-CAM, illuminate the model-specific regions of input data that contribute most to predictions, fostering an understanding of where the model focuses its attention~\cite{selvaraju2017grad}.
Furthermore, in medical domain, decision understanding is often achieved through interactive dashboards that visualise model outcomes and insights, allowing end-users to assess predictions, contributing factors, and uncertainties for informed decision-making.
These concepts collectively illuminate the intricate inner workings of machine learning models, promoting transparency and user confidence.

This article explores recent advancements in Vision Transformers for medical image analysis and presents a concise review of interpretative approaches that understand the decision-making process of ViT models. Through a detailed analysis of state-of-the-art techniques, we aim to pave the way for transparent and reliable AI systems beneficial for medical diagnosis, ultimately improving patient outcomes and healthcare practices.
In the following, we delve into general explainable AI in the medical industry and the method to quantify model output. Subsequently, we provide preliminaries on Vision Transformers, followed by the latest state-of-the-art approaches for interpreting and visualising the model output. Finally, we discuss the current direction of interpretable medical models and explore potential future directions.

\section{Interpretable Model: e\textbf{X}plainable \textbf{AI} (XAI)}

In recent years, AI has witnessed remarkable advancements, showcasing its potential to transform industries and reshape the way we interact with technology. From autonomous vehicles to medical diagnostics, AI systems are increasingly being integrated into various domains to make complex decisions and predictions.
However, as AI applications become more ubiquitous, concerns have arisen regarding their transparency, accountability, and trustworthiness~\cite{shin2020user}.
When AI model is used in mission critical applications like medical diagnosis~\cite{balasubramaniam2022transparency}, Cybersecurity~\cite{lai2023ensemble} or autonomous driving~\cite{imai2019legal}, transparency enables stakeholders to comprehend the rationale behind the model's predictions, ensuring accountability, identifying potential biases, and facilitating trust in high-stakes scenarios.

Transparency in Machine Learning (ML), also known as interpretability or explainability, aims to uncover the inner workings of intricate models~\cite{gilpin2018explaining}. From a human-centred design standpoint, transparency is not an inherent property of the ML model but rather a relationship between the algorithm and its users. Therefore, achieving transparency requires prototyping and user evaluations to develop solutions that promote transparency~\cite{balasubramaniam2022transparency}.
In specialised and high-stakes domains like medical image analysis, adhering to human-centred design principles proves challenging due to restricted access to end users and the knowledge disparity between users and ML designers.

The concept of e\textbf{X}plainable \textbf{AI} (XAI)~\cite{arrieta2020explainable} has emerged as a crucial research area that seeks to shed light on the inner workings of AI models and elucidate the factors contributing to their decisions. By providing interpretable insights into the decision-making process, XAI bridges the gap between the complexity of deep learning models and the human understanding required in critical domains like medical imaging.
In the context of medical imaging, XAI holds immense potential to revolutionise how medical professionals interact with AI systems~\cite{banerjee2020medical}. By uncovering the underlying rationales behind AI-generated diagnoses and highlighting the relevant features driving the decisions, XAI enhances medical practitioners' diagnostic accuracy and confidence and enables them to make informed and responsible clinical decisions~\cite{loh2022application}.
Ultimately, this can provide more confidence for medical professionals and patients to entrust the diagnosis recommendation and guide their decisions.
Moreover, interpretable AI models facilitate the identification and rectification of biases, enabling fair and unbiased decision-making.

XAI methods allow end users, such as clinicians, to understand, verify, and troubleshoot the decisions made by these systems~\cite{8419428}. The interpretability of ML models is crucial for ensuring accountability and trust from physicians and patients. For instance, a model detecting pneumonia is less likely to be trusted if it cannot explain why a patient received that diagnosis. In contrast, a model that provides insights into its reasoning is more likely to be appreciated and accepted. Interpretable ML systems offer explanations that enable users to assess the reliability of forecasts and recommendations, helping them make informed decisions based on the underlying logic. Moreover, addressing the potential biases in machine learning systems is essential to ensure fair ratings for individuals of all racial and socioeconomic backgrounds~\cite{doi:10.1177/2053951715622512}. The widespread use of predictive algorithms, as seen in streaming services and social networks, has raised concerns about their societal impact, including the deskilling of professionals like doctors~\cite{arnold2021teasing}. Therefore, while the application of machine learning techniques in healthcare is inevitable, establishing standardised criteria for interpretable ML in this field is urgently needed to enhance transparency, fairness, and safety.

\section{Primaries on Vision Transformer}

Vision Transformer (ViT)~\cite{dosovitskiy2020image} is a deep learning model that has gained significant attention in computer vision.
In contrast to traditional convolutional neural networks (CNNs), which have been the dominant architecture for image recognition tasks, ViT adopts a transformer-based architecture inspired by its success in natural language processing (NLP)~\cite{vaswani2017attention}. ViT breaks down an image into fixed-size patches, which are then linearly embedded and processed using a transformer encoder to capture global contextual information. This approach allows ViT to handle local and global image features effectively, leading to remarkable performance in various computer vision tasks, including image classification and object detection.

\begin{figure}[tb]
	\centering
	\includegraphics[width=1.0\linewidth]{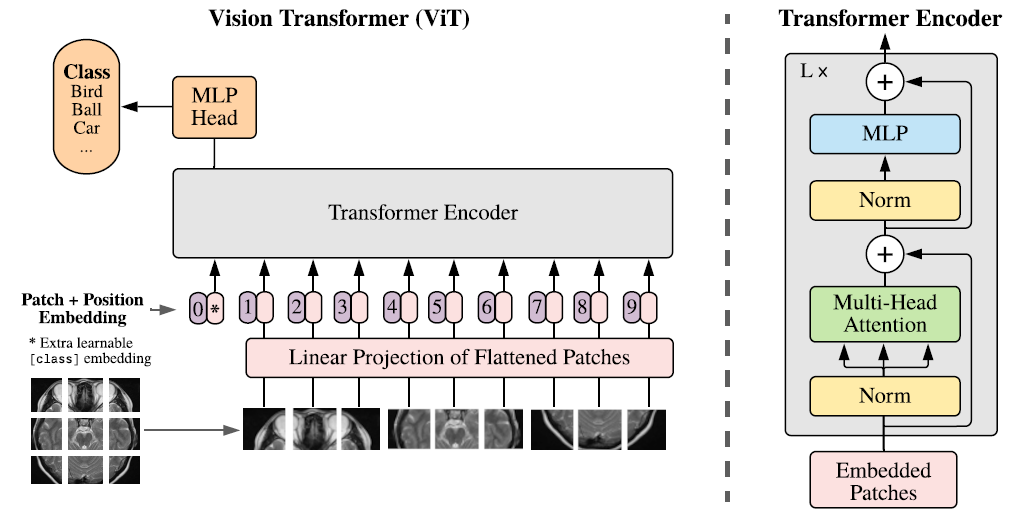}
	\caption{The basic framework of Vision Transformer (ViT) \cite{dosovitskiy2020image} and its encoder architecture.}
	\label{ViT}
\end{figure}

Generally, a vision Transformer consists of a patch embedding layer and several consecutively connected encoders, as depicted in~\Cref{ViT}.
The self-attention layer is the key component that enables ViT to achieve many state-of-the-art vision recognition performances.
The self-attention layer first transforms the input image into three different vectors---the query vector, the key vector, and the value vector.
Subsequently, the attention layer then computes the scores between each pair of vectors and determines the degree of attention when given other tokens.

Formally, given an image $\mathit{x} \in \mathbb{R}^{H \times W \times 3}$, the patch embedding layer first splits and flattens the sample $\mathit{x}$ into sequential patches $\mathit{x}_p \in \mathbb{R}^{N \times (P^2 d)}$, where $\mathit{(H,W)}$ represents the \emph{height} and \emph{width} of the input image, $(P, P)$ is the resolution of each image patch, $\mathit{d}$ denotes the output channel, and $N = {HW}/{P^2}$ is the number of image tokens.
The list of patch tokens is further fed into Transformer encoders for attention calculation.

Each Transformer encoder mainly consists of two types of sub-layers---a multi-head self-attention layer~(MSA) and an MLP layer.
In MSA, the tokens are linearly projected and further re-formulated into three vectors, namely $\bm{Q}, \bm{K}$ and $\bm{V}$. The self-attention calculation is performed on $\bm{Q}, \bm{K}$ and $\bm{V}$ by
\begin{equation}
    x'_\ell = {\rm Attention}(\bm{Q}, \bm{K}, \bm{V}) = {\rm Softmax}(\frac{\bm{Q} \cdot \bm{K}^\top}{\sqrt{d}})\cdot\bm{V},
    \label{eq:attn}
\end{equation}
where $x'_\ell$ are the tokens produced by MSA at the $\ell$-th layer.
We can formulate self-attention as
The output tokens  $x'_\ell$ are further normalised with Layer Normalisation (LN) and sent to an MLP block which consists of two fully connected layers with a GELU activation~\cite{hendrycks2016gaussian} in between. This process is formally formulated as follows,
\begin{equation}
    \mathit{x}_\ell = {\rm MLP}({\rm LN}(x'_\ell)) + x'_\ell,
    \label{eq:ffn}
\end{equation}
where $x_\ell$ is the output of the $\ell$-th encoder block.
At the last transformer layer, the high-dimensional embeddings are used for various downstream tasks, e.g., using the embeddings for training an object recognition model.

\section{Explainability Methods in XAI}

The importance of interpretability in machine learning models is widely acknowledged, but defining what constitutes interpretability remains a challenge~\cite{Lipton2016mythos}. Various definitions have been proposed, emphasising openness, accuracy, reliability, and understandability~\cite{Lipton2016mythos, Freitas2013comprehensible}. However, these definitions often overlook the user's perspective, and their needs are not adequately addressed in the produced explanations~\cite{Miller17a}. This is especially relevant in interpretable machine learning systems, where the audience's understanding and trust in the models are crucial.

Interpretability becomes even more critical in medical imaging as it influences clinicians' decision-making and patients' acceptance of the model's predictions. Interpretable machine learning systems offer valuable insights into their reasoning, helping users, such as clinicians, comprehend and verify predictions, ensuring fairness and unbiased outcomes for diverse populations. As deep learning algorithms find numerous applications in healthcare, the demand for interpretable models grows, necessitating the establishment of uniform criteria for interpretable ML in this vital domain.
The following summarises explainability methods that are commonly used in the XAI field.

\subsection{Gradient-weighted Class Activation Mapping (Grad-CAM) Method}

Grad-CAM is a gradient-based interpretability technique introduced by Selvaraju et al.~\cite{selvaraju2017grad} that aims to generate a localisation map of the significant regions in an image that contribute the most to the decision made by a neural network. Leveraging the spatial information retained in convolutional layers, Grad-CAM utilises the gradients propagated to the last convolutional layer to attribute importance values to each network neuron with respect to the final decision. An appealing advantage of Grad-CAM over similar methods is its applicability without requiring re-training or architectural changes, making it readily adaptable to various CNN-based models. Moreover, combined with Guided Backpropagation through element-wise multiplication, known as Guided Grad-CAM, it enables the generation of high-resolution and class-discriminative visualizations~\cite{selvaraju2017grad}.

\subsubsection{Saliency Maps}

Saliency Maps, introduced by Simonyan et al.~\cite{simonyan2013deep}, is a gradient-based visualisation technique that sheds light on the contribution of individual pixels in an image to its final classification made by a neural network. This method involves a backward pass through the network to calculate the gradients of the loss function with respect to the input's pixels~\cite{molnar2020interpretable}. Doing so reveals the impact of each pixel during the backpropagation step, providing insights into how much each pixel affects the final classification, particularly concerning a specific class of interest. The results from Saliency Maps can be interpreted as another image, either the same size as the input image or easily projectable onto it, highlighting the most important pixels that attribute the image to a specific class~\cite{simonyan2013deep}.

\subsubsection{Concept Activation Vectors (CAVs)}

Concept Activation Vectors (CAVs)~\cite{kim2017interpretability} represents an interpretability technique that offers global explanations for neural networks based on user-defined concepts~\cite{molnar2020interpretable}. To leverage CAVs, two datasets need to be gathered: one containing instances relevant to the desired concept and the other comprising unrelated images serving as a random reference. For a specific instance, a binary classifier is trained on these two datasets to classify between instances related to the concept of interest and unrelated ones. The CAV is then derived as the coefficient vector of this binary classifier. Testing with CAVs (TCAVs) allows averaging the concept-based contributions from the relevant dataset and comparing them to the contributions from the random dataset regarding the class of interest. Consequently, CAVs establish connections between high-level user-defined concepts and classes, both positively and negatively. This approach is particularly useful in the medical field, where medical specialists can conveniently relate the defined concepts with existing classes without delving into the intricacies of neural networks~\cite{kim2017interpretability}.

\subsubsection{Deep Learning Important FeaTures (DeepLift)}

Deep Learning Important FeaTures, commonly known as DeepLift~\cite{shrikumar2017learning}, is an explainability method capable of determining contribution scores by comparing the difference in neuron activation to a reference behaviour. By employing backpropagation, DeepLift quantifies the contribution of each input feature when decomposing the output prediction. By comparing the output difference between the original input and a reference input, DeepLift can assess how much an input deviates from the reference. One of the significant advantages of DeepLift is its ability to overcome issues related to gradient zeroing or discontinuities, making it less susceptible to misleading biases and capable of recognising dependencies that other methods may overlook. However, carefully considering the reference input and output is essential for achieving meaningful results using DeepLift~\cite{shrikumar2017learning}.

\subsubsection{Layer-wise Relevance Propagation (LRP)}

Layer-wise Relevance Propagation (LRP)~\cite{pone-lrp} is an explainability technique that provides transparent insights into complex neural network models, even with different input modalities like text, images, and videos. LRP propagates the prediction backward through the model, ensuring that the neurons' received relevance is equally distributed among the lower layers. The proper set of parameters and LRP rules make achieving high-quality explanations for intricate models feasible.

\subsubsection{Guided Backpropagation}

Guided Backpropagation~\cite{Guided_Backpropagation} is an explanation method that combines ReLU and deconvolution, wherein at least one of these is applied with masked negative values. By introducing a guidance signal from the higher layer to the typical backpropagation process, Guided Backpropagation prevents the backward flow of negative gradients, corresponding to the neurons that decrease visualised activation of the higher layer unit. This technique is particularly effective without switching, allowing visualisation of a neural network's intermediate and last layers.

\begin{figure}[t]
	\centering
    \begin{subfigure}[t]{.495\linewidth}
        \includegraphics[width=\linewidth]{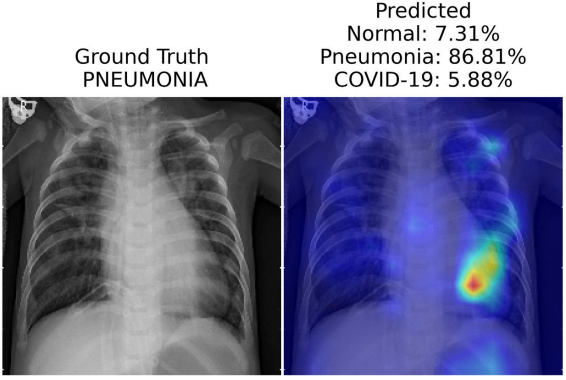}
        \caption{
            CXR of Pneumonia patient
        }
    \end{subfigure}\hfill%
    \begin{subfigure}[t]{.495\linewidth}
        \includegraphics[width=\linewidth]{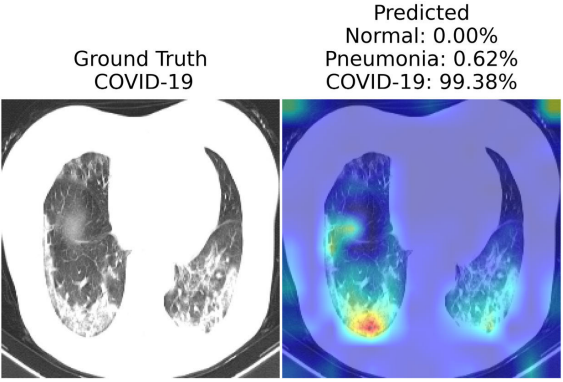}
        \caption{
            CT scan of COVID-19 patient
        }
    \end{subfigure}
	\caption{\label{fig:covid19-pred}
        Visualisation of interpreting results of xViTCOS~\cite{mondal2021xvitcos} using explainability method in~\cite{chefer2021transformer}.
        The figures highlight the associated critical factors that explain the model's decision-making.
    }
\end{figure}

\section{Vision Transformer for Medical Images}

ViTs have proven to be effective in solving a wide range of vision problems, thanks to their capability to capture long-range relationships in data. Unlike CNNs, which rely on the inductive bias of locality within their convolutional layers, vanilla ViTs directly learn these relationships from the data.
However, the success of ViTs has also brought challenges in interpreting their decision-making process, mainly due to their long-range reasoning capabilities.
In this section, we will first review some applications of ViTs in medical domains, where they are used as black box methods.
Subsequently, we will discuss other models that aim to interpret ViT's model output and provide explanations for their predictions.

\subsection{Black Box Methods}

TransMed~\cite{dai2021transmed} is a pioneering work that introduces the use of Vision Transformers (ViTs) for medical image classification. Their architecture, TransMed, combines the strengths of Convolutional Neural Networks (CNNs) for extracting low-level features and ViTs for encoding global context. TransMed focuses on classifying parotid tumours in multi-modal MRI medical images and employs a novel image fusion strategy to effectively capture mutual information from different modalities, yielding competitive results on their privately collected parotid tumour classification dataset.

\citeauthor{lu2021smile}~\cite{lu2021smile} propose a two-stage framework for glioma sub-type classification in brain images. The framework performs contrastive pre-training and then uses a transformer-based sparse attention module for feature aggregation. Their approach demonstrates its effectiveness through ablation studies on the TCGA-NSCLC~\cite{napel2014nsclc} dataset.
\citeauthor{gheflati2021vision}~\cite{gheflati2021vision} systematically evaluates pure and hybrid pre-trained ViT models for breast cancer classification. Their experiments on two breast ultrasound datasets show that ViT-based models outperform CNNs in classifying images into benign, malignant, and normal categories.

Several other works employ hybrid Transformer-CNN architectures for medical image classification in different organs. For instance, \citeauthor{khan2021gene}~\cite{khan2021gene} propose Gene-Transformer to predict lung cancer subtypes, showcasing its superiority over CNN baselines on the TCGA-NSCLC~\cite{napel2014nsclc} dataset. \citeauthor{chen2021gashis}~\cite{chen2021gashis} presents a multi-scale GasHis-Transformer for diagnosing gastric cancer in the stomach, demonstrating strong generalisation ability across other histopathological imaging datasets. \citeauthor{jiang2021method}~\cite{jiang2021method} propose a hybrid model combining convolutional and transformer layers for diagnosing acute lymphocytic leukemia, utilising a symmetric cross-entropy loss function.

\subsection{Interpretable Vision Transformer}

Interpretable vision models aim to reveal the most influential features contributing to a model's decision.
We can visualise the most influential region contributing to ViT's predictions with methods such as saliency-based techniques and Grad-CAM.
Thanks to their interpretability, these models are particularly valuable in building trust among physicians and patients, making them suitable for practical implementation in clinical settings.
\Cref{table:summary} provides a high-level overview of existing state-of-the-art interoperability methods that are specifically designed for transformer models.
Na\"ive method that only visualises the last attentive block will often be uninformative.
In addition, some interoperability methods might be class-agnostic, which means the visualisation remains the same for the prediction of all classes (e.g. rollout~\cite{abnar2020quantifying}).
In contrast, some correlation methods can illustrate different interpretation results for different target classification results (e.g. transformer attribution~\cite{chefer2021transformer}).

ViT-based methods can be used for COVID-19 diagnosis~\cite{park2021vision}, where the low-level CXR features can be extracted from a pre-trained self-supervised backbone network.
SimCLR~\cite{chen2020simple} is a popular backbone using contrastive-learning-based model training methods.
The backbone network extracts abnormal CXR feature embeddings from the CheXpert dataset \cite{irvin2019chexpert}.
The ViT model then uses these embeddings for high-level COVID-19 diagnosis. Extensive experiments on three CXR test datasets from different hospitals show their approach's superiority over CNN-based models. They also validate the generalisation ability of their method and use saliency map visualisations \cite{chefer2021transformer} for interpretability.
Similarly, COVID-ViT~\cite{gao2021covid} is another ViT-based model for classifying COVID from non-COVID images in the MIA-COVID19 challenge \cite{kollias2021mia}. Their experiments on 3D CT lung images demonstrate the ViT-based approach's superiority over the DenseNet baseline \cite{huang2017densely} in terms of F1 score.

In another work, \citeauthor{mondal2021xvitcos}~\cite{mondal2021xvitcos} introduce xViTCOS for COVID-19 screening from lungs CT and X-ray images (see \Cref{fig:covid19-pred}). The xViTCOS model is first pre-trained on ImageNet to learn generic image representations, which is then further fine-tuned on a large chest radiographic dataset. Additionally, xViTCOS employs an explainability-driven saliency-based approach \cite{chefer2021transformer} with clinically interpretable visualisations to highlight critical factors in the predictions. The model is experimentally evaluated on the COVID CT-2A dataset \cite{gunraj2021covid} for Chest X-ray, which is effective in identifying abnormal cases.

\begin{figure}[tb]
	\centering
	\includegraphics[width=\linewidth]{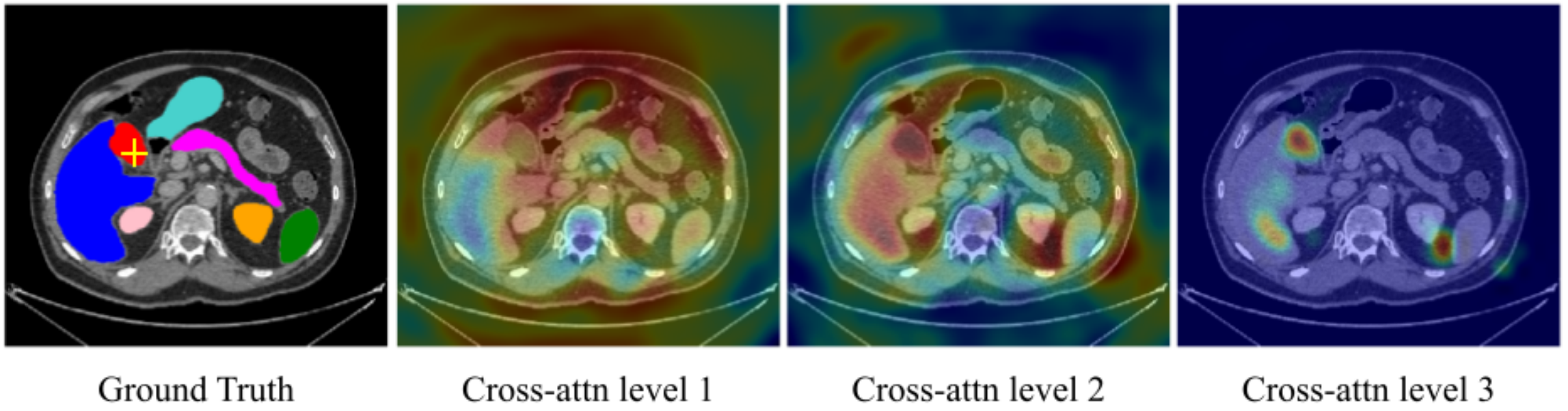}
	\caption{\label{fig:u-transformer}
        Cross-attention maps with U-Transformer~\cite{petit2021u} for the yellow-crossed pixel (left image).
        The attention maps at each level highlight the different regions contributing to the segmentation.
        \texttt{Cross-attention Level 1} is an earlier layer focusing on a wide image region.
        In contrast, We can see that \texttt{Cross-attention Level 3}, which is closer to the model output, corresponds to high-resolution feature maps and focuses on more specific regions that explain its predictions.
    }
\end{figure}

{

    \ifdefined\mytablefootnotecite
        \newcommand{\longtableinterpmethodname}{Approach}
    \else
        \newcommand{\longtableinterpmethodname}{Interoperability Method}
        \newcommand{\mytablefootnotecite}[1]{\cite{#1}}
        \newcommand{\printmytablefootnoteContent}{}
    \fi

    \begin{table}[t!]
    \renewcommand{\arraystretch}{1.4}%
        \begin{adjustwidth}{-\extralength}{0cm}
            \newcolumntype{C}{>{\centering\arraybackslash}X}
\begin{tabularx}{\linewidth}{@{}lcccccX@{}}
\toprule
\multicolumn{1}{c}{\multirow{2}{*}{\textbf{\longtableinterpmethodname}}} &
  \multicolumn{1}{c}{\multirow{2}{*}{\textbf{\begin{tabular}[c]{@{}c@{}}Class\\ Specific?\end{tabular}}}} &
  \multicolumn{4}{c}{\textbf{Metrics}} &
  \multicolumn{1}{c}{\multirow{2}{*}{\textbf{Highlights and Summary}}} \\ \cmidrule(lr){3-6}
\multicolumn{1}{c}{} &
  \multicolumn{1}{c}{} &
  \multicolumn{1}{c}{\textbf{Pixel Acc.}} &
  \multicolumn{1}{c}{\textbf{mAP}} &
  \multicolumn{1}{c}{\textbf{mF1}} &
  \multicolumn{1}{c}{\textbf{mIoU}} &
  \multicolumn{1}{c}{} \\ \midrule
Raw   Attention &  & 67.87           & 80.24 & 29.44 & 46.37 & Raw attention only consider the attention map of the last block of the transformer architecture        \\
Rollout~\mytablefootnotecite{abnar2020quantifying} &
   &
  73.54 &
  84.76 &
  43.68 &
  55.42 &
  Rollout assume a linear   Combination of tokens and quantify the influence of skip connections with   identity mateix \\
GradCAM~\mytablefootnotecite{selvaraju2017grad}         &  \checkmark & 65.91           & 71.60  & 19.42 & 41.30  & Provides a class-specific   explanation by adding weights to gradient based feature map \\
Partial   LRP~\mytablefootnotecite{voita2019analyzing} &
   &
  76.31 &
  84.67 &
  38.82 &
  57.95 &
  Considers the information flow   within the network  by identifying the   most important heads in each encoder layer through relevance propagation \\
 Transformer Attribution~\mytablefootnotecite{chefer2021transformer} & \checkmark
   &
  76.30 &
  85.28 &
  41.85 &
  58.34 &
  Combines relevancy and attention-map gradient by regarding the gradient   as a weight to the relevance for certain prediction task \\
Generic Attribution~\mytablefootnotecite{chefer2021generic} & \checkmark
   &
  79.68 &
  85.99 &
  40.10 &
  61.92 &
  Generic attribution extends the   usage of Transformer attribution to co-attention and self-attention based   models with a  generic relevancy update   rule \\
Token-wise Approx.~\mytablefootnotecite{chen2022beyond}   {\gdef\lockmytablefootnotecite{}} & \checkmark
   &
  82.15 &
  88.04 &
  45.72 &
  66.32 &
  Uses head-wise and token-wise approximations to visualise tokens interaction in the pooled vector with   noise-decreasing strategy \\ \bottomrule
\end{tabularx}
        \end{adjustwidth}
    \caption{Summary table on interoperability approaches for transformer models.
    \emph{Class-specific} refers to whether the approach can attribute different attentive scores that are specific to the predicted class (in multi-class predictions).
    \emph{Metrics} used to evaluate each methods are \emph{Pixel Accuracy}, \emph{mean Average Precision (mAP)}, \emph{mean F1 score (mF1)}, and \emph{mean Intersection over Union (mIoU)}.
    \label{table:summary}}
    {\printmytablefootnoteContent{}}
    \end{table}

}

\citeauthor{shome2021covid}~\cite{shome2021covid} have introduced another ViT-based model for the diagnosis of COVID-19 infection at a large scale. They combined multiple open-source COVID-19 CXR datasets to accomplish this, forming a comprehensive multi-class and binary classification dataset. In order to enhance visual representation and model interpretability, they implemented Grad-CAM-based visualization~\cite{selvaraju2017grad}.
The Transformer-based Multiple Instance Learning (TransMIL) architectures proposed by Shao \textit{et al.} \cite{shao2021transmil} aims to address whole slide brain tumour classification. Their approach involves embedding patches from whole slide images (WSI) into the feature space of a ResNet-50 model. Subsequently, the sequence of embedded features undergoes a series of processing steps in their proposed pipeline, including squaring the sequence, correlation modelling, conditional position encoding using the Pyramid Position Encoding Generator (PPEG) module, local information fusion, feature aggregation, and mapping from the transformer space to the label space. This innovative approach holds promise for accurate brain tumour classification, as illustrated in their work \cite{shao2021transmil}.
The self-attention module in transformers can leverage global interactions between encoder features, while cross-attention in the skip connections allows a fine spatial recovery.
For example,~\Cref{fig:u-transformer} highlights the attention level across the whole image for a segmentation task in a U-Net Transformer architecture~\cite{petit2021u,huang2022swin}.

In whole slide imaging (WSI) based pathology diagnosis, annotating individual instances can be expensive and laborious. Therefore, a label is assigned to a set of instances known as a "bag." This weakly supervised learning type is called Multiple Instance Learning (MIL)~\cite{fung2007multiple}, where a bag is labelled positive if at least one instance is positive or negative when all instances in a bag are negative. However, most current MIL methods assume that the instances in each bag are independent and identically distributed, overlooking any correlations among different instances.

\begin{figure}[tb]
	\centering
	\includegraphics[width=\linewidth]{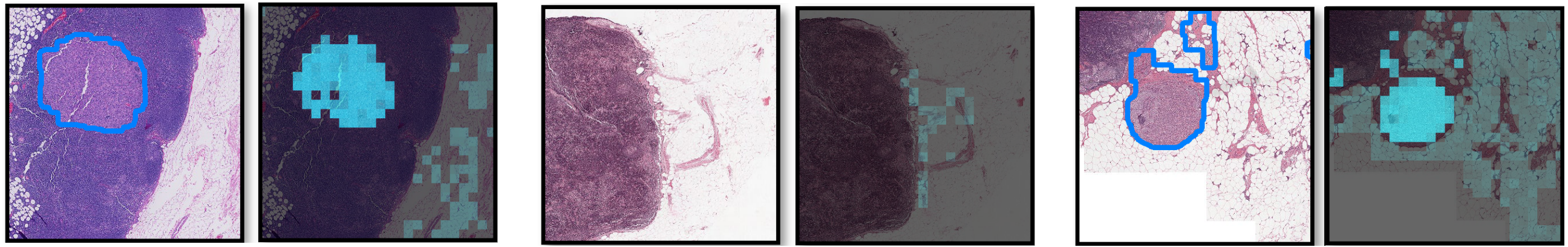}
	\caption{\label{fig:cancer-atten}
        Visualise the probability derivation output from~\cite{zhang2022dtfd} lung cancer region detection.
        Each pair of images contains (left) ground truth with the tumour regions delineated by blue lines and (right) the probability derivation output.
        Brighter cyan colours indicate higher probabilities of being tumours for the corresponding locations.
        We can see that most high cyan region localises the positive detection regions.
    }
\end{figure}

To address this limitation, \citeauthor{shao2021transmil}~\cite{shao2021transmil} proposes TransMIL, a novel approach that explores morphological and spatial information in weakly supervised WSI classification. Their method aggregates morphological information using two transformer-based modules and a position encoding layer. To encode spatial information, they introduce a pyramid position encoding generator.
TransMIL achieves state-of-the-art performance on three computational pathology datasets: CAMELYON16 (breast)~\cite{bejnordi2017diagnostic}, TCGA-NSCLC (lung)~\cite{napel2014nsclc}, and TCGA-R (kidney). Their approach demonstrates superior performance and faster convergence than CNN-based state-of-the-art methods, making it a promising and interpretable solution for histopathology classification.
Attention-based ViT can further derive instance probability for highlighting regions of interest.
For example, AB-MIL~\cite{zhang2022dtfd} uses the derivation of instance probability for feature distillation as shown in~\Cref{fig:cancer-atten}.
The attentive method can also be used for interpreting the classification of retinal images~\cite{playout2022focused}.

For the diagnosis of lung tumours, Zheng \citeauthor{zheng2021deep}~\cite{zheng2021deep} proposes the graph transformer network (GTN), leveraging the graph-based representation of WSI. GTN consists of a graph convolutional layer~\cite{kipf2016semi}, a transformer, and a pooling layer. Additionally, GTN utilises GraphCAM~\cite{chefer2021transformer} to identify regions highly associated with the class label. Thorough evaluations on the TCGA dataset~\cite{napel2014nsclc} demonstrate the effectiveness of GTN in accurately diagnosing lung tumours. This graph-based approach provides valuable insights into the spatial relationships among regions, enhancing the interpretability of the classification results.

\section{Conclusion}

In this review, we explored the advancements and applications of interpretability techniques in the context of deep learning algorithms applied to various domains, with a particular focus on the medical field. Interpretable machine learning models have become increasingly essential as complex models like Transformers and Vision Transformers (ViTs) gain popularity due to their impressive performance in various tasks.

We discussed several interpretability methods, such as Grad-CAM and Saliency Maps, which provide insights into the decision-making process of deep learning models. These methods allow users, including medical practitioners and researchers, to understand and trust the predictions made by these models. Moreover, they enable the identification of biases, contributing factors, and regions of importance, leading to more informed and accountable decision-making.
Medical images play a crucial role in clinical diagnostics, offering valuable insights into various anatomical areas. Deep learning algorithms have shown exceptional capabilities in healthcare applications, including medical image analysis and patient risk prediction. However, interpretability is particularly important in medical image analysis as it empowers end-users, such as clinicians and patients, to understand and trust the model's decisions. Interpretable machine learning models provide explanations behind their predictions, allowing users to assess and validate the output before making critical decisions. This transparency is essential for ensuring fairness, mitigating biases, and building trust in machine learning systems, especially in healthcare. By fostering transparency and accountability, interpretable ML in medical imaging has the potential to revolutionise patient care and enhance healthcare practices globally.

In the medical domain, interpretable deep learning models have shown promising results in various applications, such as COVID-19 diagnosis, brain tumour classification, breast cancer detection, and lung tumour subtyping. By visualising the learned features and attention scores, these models achieve high accuracy and provide valuable insights into disease detection and classification, ultimately improving patient care and treatment decisions.

\section{Limitation and Future Directions}

There are several limitations that ViTs might exhibit when applied in the medical imagery domain.
While the global context understanding is one of the strengths of self-attentive ViTs, it can also become a limitation.
In certain tasks where local features are crucial, ViTs might struggle to focus on specific localised patterns, potentially leading to suboptimal performance.
Therefore, ViTs are often paired with convolutional layers to account for feature extraction in the local-scale~\cite{xie2021cotr}.
Self-attentive ViTs also often require large amounts of training data to generalise well, which is a limitation in the medical domain as collecting large annotated datasets is challenging.
In addition, self-attention mechanisms, especially when applied to large input images, can be computationally expensive. The quadratic complexity of self-attention hampers its scalability to high-resolution images, making it less efficient compared to convolutional neural networks (CNNs) for some tasks.
Training large ViT models with medical images can often be a practical difficulty due to the computational resources that it required when paired with high-resolution medical images.
As a result, small clinics without access to high-performing computational resources might need to rely on external resources to train or infer ViT models.

\begin{figure}[tb]
	\centering
	\includegraphics[width=1.0\linewidth]{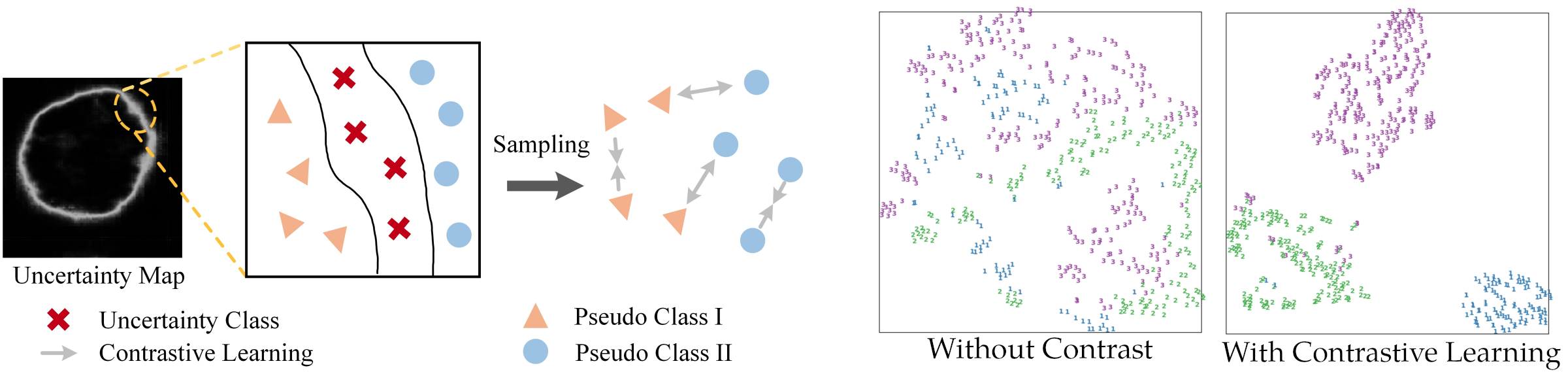}
	\caption{\label{fig:contrastive-learning}
        Visualising results of incorporating contrastive learning in using CNN-transformer decoder~\cite{wang2022uncertainty}.
        (Left) Using uncertainty region as a guidance to implements contrastive learning in medical images in an unsupervised manner.
        (Right) Visualising the separation of predicted categories, with and without contrastive learning (dimensional reduction using t-SNE algorithm; colours of markers represent pixel categories).
    }
\end{figure}

While interpretable machine learning has made significant progress, several exciting avenues exist for future research and development.
For example, developing unified frameworks that \textbf{combine multiple interpretability techniques} can provide a more comprehensive understanding of model decisions. These frameworks should be scalable and adaptable to various deep learning architectures, enabling users to choose the most suitable methods for their specific applications.
The uncertainty within the dataset should also be quantifiable.
\textbf{Integrating uncertainty estimation} into interpretable models can enhance their reliability and robustness. Uncertainty quantification can be crucial, especially in critical medical decision-making scenarios, where a model's confidence can significantly impact patient outcomes.
\Cref{fig:contrastive-learning} illustrates an example of using uncertainty estimation for improving the medical image segmentation with a transformer-based model.
In particular, the proposed model in~\cite{wang2022uncertainty} takes medical images as input and generates an uncertainty map representation in an unsupervised manner.
Then, the uncertain region in the uncertainty map is used to reduce the possibility of noise sampling, and to encourage consistent network outputs.
Such an approach can then predict classes with a greater degree of separation (see right of~\cref{fig:contrastive-learning}), which in turn improve the transparency for medical diagnosis with less miss-classification.
A similar approach that addresses uncertainty within the dataset can often encourage models to produce grounded predictions.
By addressing these future directions, we can enhance the trust, transparency, and effectiveness of deep learning models in medical applications, ultimately improving patient outcomes and healthcare practices.

\printbibliography

\end{document}